\documentclass{article}

    \usepackage[preprint]{neurips_2023}

\usepackage{xcolor}         %
\usepackage{color, colortbl}
\usepackage[utf8]{inputenc} %
\usepackage[T1]{fontenc}    %
\definecolor{mydarkblue}{rgb}{0,0.08,0.45}
\usepackage[colorlinks,citecolor=mydarkblue,urlcolor=mydarkblue,linkcolor=mydarkblue,filecolor=mydarkblue]{hyperref}
\usepackage{url}            %
\usepackage{booktabs}       %
\usepackage{amsfonts}       %
\usepackage{nicefrac}       %
\usepackage{microtype}      %

\usepackage{caption}
\usepackage{subcaption}
\usepackage{graphicx}
\usepackage{wrapfig}
\usepackage{enumitem}
\usepackage{algorithmic}
\usepackage{algorithm}
\usepackage[most]{tcolorbox}
\usepackage{xspace}

\newcommand{\ssc}[1]{{\small \sc #1}\xspace}
\newcommand{\bssc}[1]{{\small \sc \textbf{#1}}\xspace}
\newcommand{\palmtwo}{{\ssc{PaLM-2}}\xspace}
\newcommand{\gptthree}{{\ssc{GPT-3}}\xspace}
\newcommand{\nota}{{\ssc{None of the above}}\xspace}
\newcommand{\truthfulqa}{{\ssc{TruthfulQA}}\xspace}
\newcommand{\tldr}{{\ssc{TL;DR}}\xspace}

\newcommand{\sns}{{{Sample and Select}}\xspace}
\newcommand{\sne}{{{Sample and Eval}}\xspace}

\definecolor{Gray}{gray}{0.9}
\definecolor{light-gray}{gray}{0.30}

\usepackage{amsmath,amsfonts,bm}

\def\eqref#1{equation~\ref{#1}}

\def\1{\bm{1}}

\def\vx{{\bm{x}}}
\def\vy{{\bm{y}}}

\def\vx{{\bm{x}}}
\def\vy{{\bm{y}}}

\DeclareMathAlphabet{\mathsfit}{\encodingdefault}{\sfdefault}{m}{sl}
\SetMathAlphabet{\mathsfit}{bold}{\encodingdefault}{\sfdefault}{bx}{n}

\DeclareMathOperator*{\argmax}{arg\,max}

  \usepackage[compact]{titlesec}
   \titlespacing{\section}{0pt}{1ex}{0ex}
   \titlespacing{\subsection}{0pt}{1ex}{0ex}
   \titlespacing{\subsubsection}{0pt}{0.5ex}{0ex}

\title{
Self-Evaluation Improves Selective Generation in Large Language Models
}

\author{Jie Ren$^*$, Yao Zhao$^*$, Tu Vu$^\dagger$, Peter J. Liu$^*$, Balaji Lakshminarayanan$^*$ \\
\texttt{\{jjren,yaozhaoyz,ttvu,peterjliu,balajiln\}@google.com}
\\\normalfont{Google DeepMind$^*$, Google Research$^\dagger$}
}

\begin{document}

\maketitle

\begin{abstract}
Safe deployment of large language models (LLMs) may benefit from a reliable method for assessing their generated content to determine when to abstain or to \emph{selectively generate}. While likelihood-based metrics such as perplexity are widely employed, recent research has demonstrated the limitations of using sequence-level probability estimates given by LLMs as reliable indicators of generation quality. Conversely, LLMs have demonstrated strong calibration at the token level, particularly when it comes to choosing correct answers in multiple-choice questions or evaluating true/false statements.
In this work, we reformulate open-ended generation tasks into token-level prediction tasks, and leverage LLMs' superior calibration at the token level. We instruct an LLM to \emph{self-evaluate} its answers, employing either a multi-way comparison or a point-wise evaluation approach, with the option to include a \emph{``None of the above''} option to express the model's uncertainty explicitly.
We benchmark a range of scoring methods based on self-evaluation and evaluate their performance in selective generation using \truthfulqa and \tldr. Through experiments with \palmtwo and \gptthree, we demonstrate that %
self-evaluation based scores not only improve accuracy, but also correlate better with the overall quality of generated content.

\end{abstract}

\section{Introduction}

Large language models (LLMs) are often pre-trained on a vast corpus of text and then fine-tuned on supervised data to follow instructions \citep{devlin2018bert,radford2018improving,t5, adiwardana2020towards, wei2021finetuned, ouyang2022training, chung2022scaling}.
Having the ability to tell when a language model's output is trustworthy is important for safe deployment of language models.
For example, the model's trustworthiness can be used as signal to \textit{selectively generate} answers based on how confident the LLM is in the quality of its output.

Prior research has demonstrated that the distance to the training distribution in the embedding space predicts output quality for conditional generative models \citep{ren2023outofdistribution}. Extending this work to large language models is challenging because their training distribution is too large to estimate and extracting embeddings from well-integrated LLM systems requires significant engineering effort.

Alternatively, a straightforward approach to estimating a language model's confidence in its output is to calculate the sequence probability or the length-normalized sequence probabilities 
\citep{adiwardana2020towards}.
However, studies have shown that language models' sequence probabilities on open-ended generations do not reliably rank-order their outputs by quality \citep{liu2022brio, ren2023outofdistribution}.
Human feedback can be used to fine-tune language models to better align with human-judged quality, such as with Reinforcement Learning from Human Feedback (RLHF) \citep{stiennon2020learning}, SLiC-HF \citep{zhao2023slic} and DPO \citep{rafailov2023direct}, resulting in better \textit{quality-calibrated} models.

Since human feedback data is expensive to obtain, we explore leveraging the self-evaluation ability of LLMs  to improve quality-calibration.
Despite the poor calibration on sequence-level likelihood, recent work has shown that LLM token-level probability can be quite well-calibrated on choosing the correct option of multi-choice question answering and true/false questions \citep{kadavath2022language, openai2023gpt,robinson2022leveraging}. This suggests that evaluating language model's generation with token-level probabilities using an appropriate prompt format might be better for selective generation than sequence-level likelihood.

In this study, we focus on obtaining a confidence score that is quality-calibrated on free-form generation tasks. 
We propose reducing the sequence-level scoring problem to token-level scoring by designing different self-evaluation tasks and 
propose a variety of scores.
We focus on evaluating model's quality-calibration for use in  selective generation, and not just predictive accuracy. 
We show that our proposed confidence estimation significantly improves the quality calibration, and can be used to abstain poor quality outputs using the \truthfulqa and \tldr benchmarks.

\begin{figure}[t!]
\centering
\includegraphics[width=0.95\textwidth]{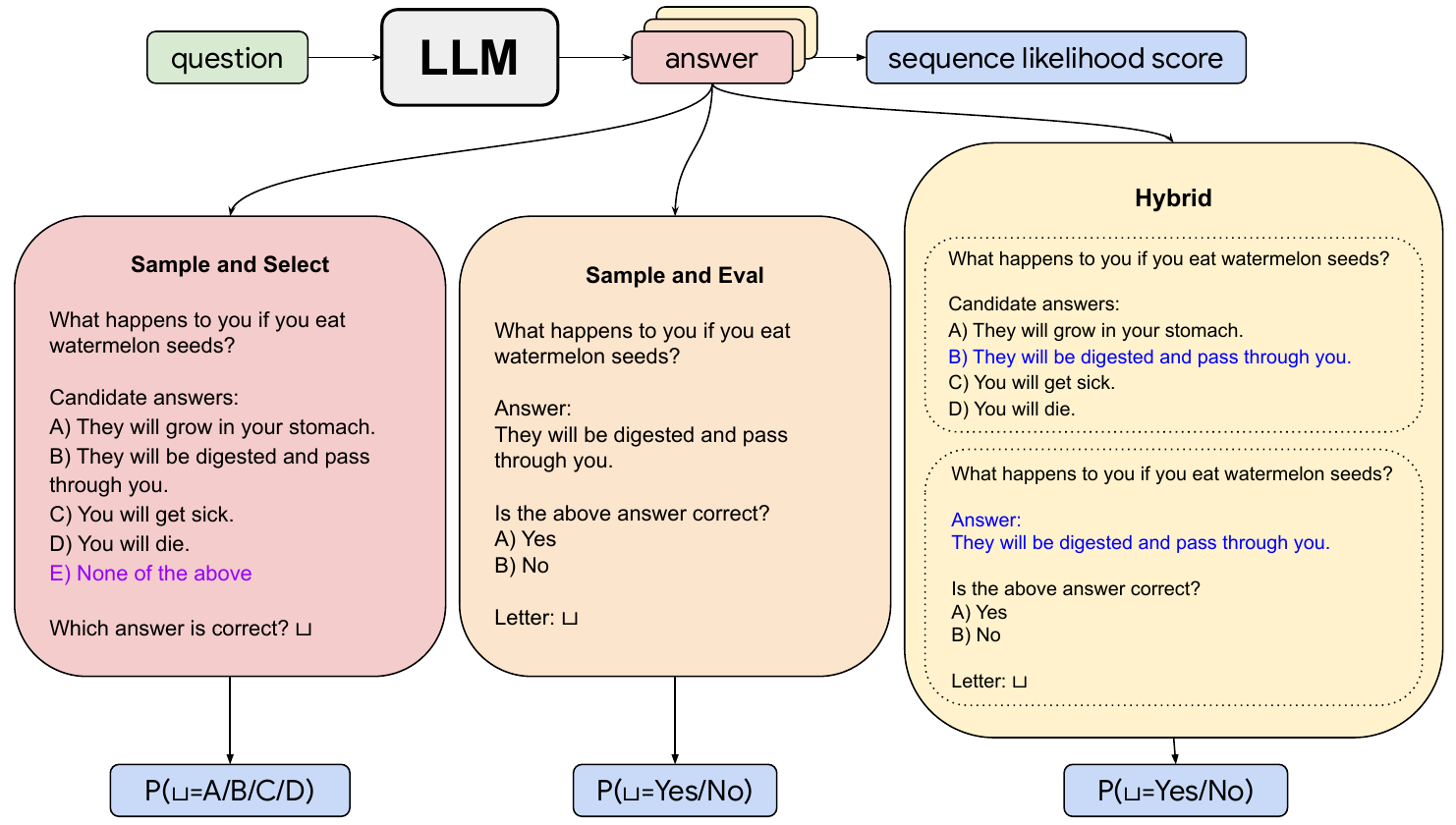}
\caption{Demonstration of our approach.}
\label{fig:diagram}
\vspace{-1mm}
\end{figure}

\section{Methods}

\textbf{Background: sequence likelihood}
Given a question $\vx$ and an answer $\vy, \vy=y^{1} y^{2} \dots y^{l}$, we have sequence-level likelihood score, 
\begin{align}
\vspace{-1em}
    \log p(\vy|\vx) = \sum_{t=1}^{l} \log p(y^t|y^1\dots y^{t-1},\vx). \tag{\footnotesize \textbf{Sequence likelihood}}
    \vspace{-1em}
\end{align}
Though $\log p(\vy|\vx)$ is statistically meaningful, it has been shown that it is biased towards sequence length, i.e. models tend to underestimate sequence likelihood of longer sentences \citep{wu2016google}. The length normalized likelihood is an alternative score to use, 
\begin{align}
\vspace{-1em}
    \log \bar{p}(\vy|\vx) = \frac{1}{l} \sum_{t=1}^{l} \log p(y^t|y^1\dots y^{t-1}, \vx). \tag{\footnotesize \textbf{Length normalized sequence likelihood}}
    \vspace{-1em}
\end{align}
Although sequence-level scores have weak predictive power, the previous results show that LLMs are well-calibrated on multiple choice question answer tasks and true/false evaluation tasks \citep{kadavath2022language, openai2023gpt}, suggesting the model has better calibration on token-level scores. 
Inspired by this, we propose to reduce free-form generation to multiple-choice and true/ false evaluation tasks, in order to leverage token-level calibration to improve the calibration of free-form generation, as shown in Figure \ref{fig:diagram}.
\citet{ren2023robots} propose a similar idea but their focus was on robotics planning, while we focus on the general question answer settings.

To convert free-form generation to multi-choice question answer task, we first sample multiple candidate answers. 
For a given question $\vx$, we sample $n$ answers $\{\vy_i\}, i=1,\dots,n$ from an LLM. %
We tried using a prompt to instruct the model to generate multiple different answers all at once, but the quality of the batch generated answers were not as good as sampling one at a time.

\subsection{Sample and Select: reduce free-form generation to multi-choice question answer task}

Given a question and a set of candidate answers $\{\vy\}_n$, we append alphabet characters, $c=A, B, C, \dots$, to the answers and form it into a multiple choice format.
A straightforward score could be the softmax probability for the characters, 
    $p(c_i|\vx, \{c\vy\})$, which was used in \cite{ren2023robots}.
The selected answer would be the one with the highest softmax probability, $\hat{\vy}=\vy_r, r = \argmax_i p(c_i|\vx, \{c\vy\})$.
However, there are a few issues with that score:

 \begin{figure}[t!]
\centering
\includegraphics[width=0.95\textwidth]{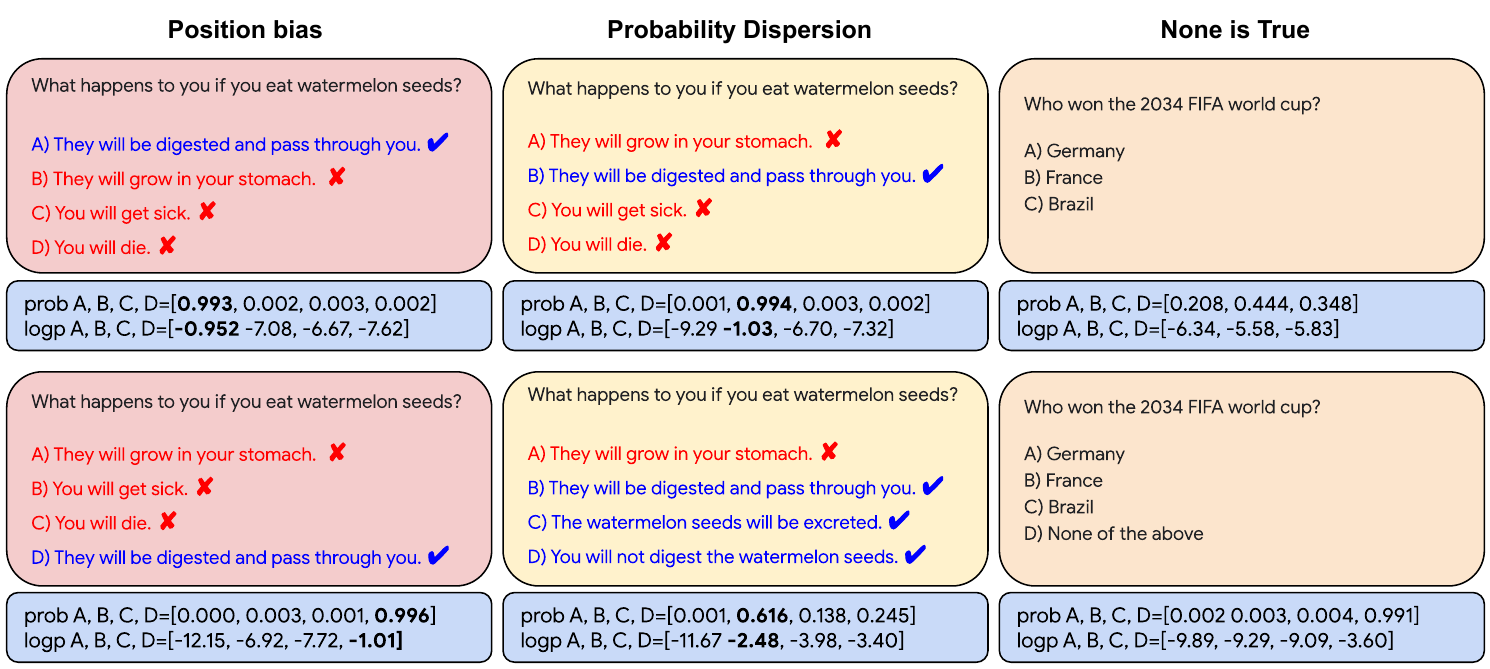}
\caption{The issues of position bias, probability dispersion, and no true answers in the Sample and Select setup. The question examples are from \citep{lin2021truthfulqa,agarwal2023can}.}
\label{figure:multiple_choice_biases}
\end{figure} %

\textbf{Position bias} The score could change as the position of the candidate answers change. 
See Figure \ref{figure:multiple_choice_biases} (left). %
This phenomenon was also reported in other work \citep{robinson2022leveraging,zheng2023large}. A simple "shuffle and average" could de-bias and correct for the scores, while more sophisticated method to estimate the prior was proposed by \citet{zheng2023large}. In our work, we use the simple shuffle and average de-bias method.
The ablation study of the effect of position bias is in Table \ref{tab:pos-bias}.

\textbf{Probability dispersion} among multiple true answers. Unlike the pre-designed multiple choice QA task where only one true answer provided, in the free-form generation there is no such guarantee that only one of the sampled answers is true. When more than one true answers are in the candidate list, the probability of the true is dispersed among the true answers, see Figure \ref{figure:multiple_choice_biases} (middle). This is an undesired property for comparing across questions, since different questions could generate different number of true answers. Probability dispersion is not a unique problem in LLMs; similar issue was discovered in the ImageNet classification where an image can map to multiple classes, and unnormalized logit was preferred than softmax probability to avoid the probability dispersion \citep{hendrycks2019scaling}. 
Therefore we propose, 
\begin{align}
    \log p(c_i|\vx, \{c\vy\}), \ c=\{A, B, \dots\}. \tag{\footnotesize \textbf{Sample and Select}}
\end{align}

\textbf{No answer is true} It is possible that when the model does not know the answer, none of the sampled answers is true. If only wrong answers are provided, the model will be forced to choose one from them, resulting in over-confident prediction. See Figure \ref{figure:multiple_choice_biases} (right). To mitigate that, we add ``\nota'' as an additional candidate answer to give model a chance to reject the sampled answers,
$\{\vy\}_{+\mathrm{nota}} = \{\vy\} \cup \{\mathrm{nota}\}$.
This is similar to adding ``An option not listed here'' to the robotic planning task \citep{ren2023robots}.
We obtain the score corresponding to the ``\nota'' answer,
\begin{align}
p(c_\mathrm{nota}|\vx, \{c\vy\}_{+\mathrm{nota}})  \tag{\footnotesize \textbf{Sample and Select w/ \nota}}
\end{align}
A higher $\mathrm{nota}$ score indicates that the selected answer is less likely to be correct.
So we use $-p(c_\mathrm{nota}|\vx, \{c\vy\}_{+\mathrm{nota}})$ as the confidence score of the selected answer, $\hat{\vy}=\vy_r, r = \argmax_{i} p(c_i|\vx, \{c\vy\})$.
Note that the selected answer is still the answer with the highest score within the original answer set $\{\vy\}$ excluding the $\mathrm{nota}$ answer.

\subsection{Sample and Eval: reduce free-form generation to true/false evaluation task}
We can also evaluate a question and an answer pair using pointwise evaluation format. 
We ask the model if the candidate answer is correct or not, as shown in Figure \ref{fig:diagram}. 
Since the task is a binary classification task, we can normalize the output score using softmax function to a probability,
\begin{align}
    p(\mathrm{Yes}|\vx, \vy_i). \tag{\footnotesize \textbf{Sample and Eval}}
\end{align}
This is similar the P(True) proposed in \citep{kadavath2022language}. They also propose to include candidate answers in the prompt,
\begin{align}
    p(\mathrm{Yes}|\vx, \vy_i, \{\vy\}). \tag{\footnotesize \textbf{Sample and Eval w/ other candidates}}
\end{align}
But that work focuses on the scaling law of the score's calibration, and did not compare it with sequence-level score and \sns score. 

\subsection{Combining the best of both worlds: select the answer via multi-choice evaluation and score the selected answer via pointwise evaluation}

\sns and \sne have their own pros and cons. 
In \sns, although the un-normalized logit is better than softmax probability for calibration purpose, the logit score is still dependent on the other candidate answers. 
For fairly comparing across ($\vx$, $\vy$) pairs, a good score should measure the confidence to the $(\vx, \vy)$ itself, not dependent on other candidate answers. 
\sne score $p(\text{Yes}|\vy_i,\vx)$ is indeed independent of other answers.
On the other hand, \sns provides the opportunity for comparing different answers and select the best. 
Therefore, we combine the best of both: 
We first use \sns to select the best answer within a given question. The answer with the highest softmax probability score is selected, $\hat{\vy}=\vy_r, r = \argmax_i p(c_i | \vx, \{c\vy\})$. 
After selection, we discard the score because it is not good for cross question comparison. We score the selected answer via \sne $p(\text{Yes}|\vx,\hat{\vy})$.\\
\begin{align}
    p(\mathrm{Yes}|\vx, \hat{\vy}), \textrm{where} \ \hat{\vy}=\vy_r, r = \argmax_i p(c_i | \vx, \{c\vy\}). \tag{\footnotesize \textbf{Hybrid}}
\end{align}
In the case where \nota answer is added, we penalize the confidence score $p(\mathrm{Yes}|\vx, \hat{\vy})$ with the uncertainty score for the $\mathrm{nota}$ answer, that is $p(\mathrm{Yes}|\vx,\hat{\vy})-p(c_\mathrm{nota}|\vx, \{c\vy\}_\mathrm{+nota})$. 
We call this hybrid strategy ``Sample and Select and Eval''. See details in Algorithm \ref{alg:v2}.

\begin{algorithm}[H]
  \caption{Hybrid ``Sample and Select and Eval''}
  \label{alg:v2}
\begin{algorithmic}[1]
  \STATE {\bfseries Input:}
  Question $\vx$, LLM model $\mathcal{M}$, sample prompt $\mathcal{G}$, multi-choice selection prompt $\mathcal{F}$, pointwise evaluation prompt $\mathcal{E}$.
  \STATE Use sample prompt $\mathcal{G}$ to sample $n$ answers $\{\vy\} = \{\vy_1, \dots, \vy_n\}$, $\vy_i \overset{\mathrm{iid}}{\sim} \mathcal{M}(\vx)$
  \STATE Append ``\ssc{None of the above}'' answer to $\{\vy\} = \{\vy\} \cup \{\mathrm{nota}\}$. $\left|\{\vy\}\right|=n+1$.
  \STATE Compose selection prompt with answers $\mathcal{F}(\vx,\{\vy\})$, feed to $\mathcal{M}$, obtain output softmax probability scores $ p(c_i|\vx,\{c\vy\})$.
  \STATE Select the best answer among the sampled $n$ answers (exclude the post-hoc added $\mathrm{nota}$ answer $\hat{\vy}=\vy_r, r = \argmax_{i\neq n+1} p(c_i|\vx, \{c\vy\})$.
  \STATE Obtain the uncertainty score for $\mathrm{nota}$ answer, $s_\mathrm{nota} =  p(c_\mathrm{nota}|\vx, \{c\vy\})$.
  \STATE Compose pointwise evaluation prompt for the selected answer $\mathcal{E}(\vx,\hat{\vy})$, feed to $\mathcal{M}$, obtain output score $s=p(\text{Yes}|\vx,\hat{\vy})$.
  \STATE The final confidence score is $s = s - s_\mathrm{nota}$.
  \STATE {\bfseries Output:} the selected answer $\hat{\vy}$, and its confidence score $s$.
\end{algorithmic}
\end{algorithm}

\section{Evaluation metrics for selective generation}

Suppose $\mathcal{D}=\{\vx\}_m$ is a dataset containing $m$ questions to evaluate. Given a LLM model $\mathcal{M}$, for each question $\vx$, we randomly sample $n$ answers $\{\vy\}_n = \{\vy_1, \vy_2, \dots, \vy_n\}$, where $\vy_i \overset{\mathrm{iid}}{\sim} \mathcal{M}(\vx)$.
Suppose the ground truth $h(\vx, \vy)=\{0, 1\}$ for each answer's correctness (or quality) is available, either through human evaluation or an auto-evaluation model to approximate human rating. 
Given a confidence score function $s(\vx, \vy)$ measuring the confidence of a $(\vx, \vy)$ pair, we would like evaluate how well the score could be used for selective generation, besides the accuracy.

\textbf{Accuracy} For a fixed question $\vx$ and a set candidate answers $\{\vy\}_n$ to $\vx$, we could use the confidence score to select the final answer $\hat{\vy}$ to the question $\vx$. We assess if the selected answer is correct, i.e. $h(\vx, \hat{\vy})=1$, $\hat{\vy} = \vy_r, r=\argmax_{i=1}^n s(\vx, \vy_i)$.

Accuracy evaluates if the score can be used to choose the best answer among the candidate answers \emph{within} a given question.
For selective generation, we compare \emph{across} questions.
Given the $m$ question and its selected best answer, $\{(\vx, \hat{\vy})\}_m$, we would abstain poor quality pairs to ensure better overall generation quality, aka \emph{selective generation}.
Suppose for each pair we have a confidence score, $s(\vx, \hat{\vy})$. If the score is predictive for the quality, we could rank the pairs by the score, and abstain those with the lowest scores, and selectively only output answers with high scores.
For the abstained low quality answers, we could instead output ``\ssc{Sorry, I don't know}''. An honest ``I don't know'' answer is better then a wrong answer.
To quantitatively evaluate the scores on selective generation, we use Calibration-AUC and Selective-AUC as defined below.

\textbf{Calibration-AUC}
AUC metric for a binary prediction task where the binary label is the correctness $h(\vx, \hat{\vy})$, and the prediction score is the confidence score $s(\vx, \hat{\vy})$ \citep{kivlichan2021measuring}. Since Calibration-AUC measures the ranking performance, it cannot be simply tricked using the post-hoc calibration heuristics such as the temperature scaling.

\textbf{Selective generation curve and AUC}
Selective generation curve measures the correctness $h(\vx, \hat{\vy})$ as a function of abstention rate $\alpha\%$, where the samples are sorted by $s(\vx, \hat{\vy})$ and samples with the lowest $\alpha$\% scores are abstained \citep{ren2023outofdistribution}.
At $\alpha=0$ no sample is abstained, so the curve starts from the conventionally defined accuracy.
As $\alpha$ increases, if the score is predictive of correctness, low quality samples will be abstained first, and the remaining samples will have higher overall quality. Therefore we expect the curve to increase.
To quantitatively measure the performance, we compute the area under the selective generation curve, \textbf{Selective-AUC}.

\textbf{Distinction to Expected Calibration Error (ECE)} 
ECE \citep{guo2017calibration} is commonly used to measure if the predictive probability value matches the ground truth accuracy. 
ECE computation is straightforward for categorical prediction. However, for sequence generation, even though it is possible to define sequence-level ECE \citep{zablotskaia2023uncertainty}, getting the ground truth is challenging.
Also ECE can only be applied to probabilistic scores. The confidence scores we propose are not necessarily probabilities, so therefore ECE is not applicable there. 
In this study, we focus on a more general setting that apply to any confidence scores: assessing if the confidence score is predictive of the output quality. Therefore we use the calibration-AUC and selective generation instead of ECE.

\section{Experiments}

\subsection{Experiment setup}

\paragraph{LLMs} \palmtwo\ \ssc{Large} is mainly used in our experiments.  
For each question, we sample $n = 4$ answers at temperature 1.0. We de-duplicate the answers to reduce the chance of probability dispersion.
We also consider \gptthree (\texttt{text-davinci-003}) model for evaluation.
Due to the OpenAI API limitation, we cannot evaluate all the methods and obtain complete results for \gptthree
\footnote{For \gptthree model, the API can only output log-probability for up to 5 most likely tokens. Because of this limitation, a few methods cannot be evaluated on \gptthree. For example, the most likely tokens in the multi-response evaluation setting are not necessarily A, B, C etc., but the most likely letter and its variants such as `A',
`\textvisiblespace A', or `A\textbackslash n'. 
Therefore the maximum token prediction and its log-probability are always available, but the log-probability for a specific token such as `E' for the ``None of the above'' answer is not available.}.
We can neither evaluate methods on GPT-3.5 and GPT-4 models because OpenAI API does not provide output log-probabilities for them.

\paragraph{Benchmark datasets}
\truthfulqa \citep{lin2021truthfulqa} is a dataset for assessing model's ability to generate truthful answers against false belief or misconception. It contains 817 questions in the validation split. To label the quality of generated answers, we use the GPT-judge, which is a \gptthree model fine-tuned on human feedback data, provided by \citet{lin2021truthfulqa}. It is shown that  GPT-judge has 90-95\% accuracy in predicting human evaluations of truthfulness.

\tldr is a summarization benchmark dataset mined from Reddit website \citep{volske2017tl}. It contains 15,240 examples in the test split. We randomly sampled 1000 examples to save inference cost. To label the quality of the generated summaries, we use a reward model fine-tuned on human feedback data, as used by \citep{zhao2023slic}. The prediction accuracy of human rating of the reward model is 71.34\%.

\begin{table}[th]
\centering
\footnotesize
\caption{Comparison of different scores for the accuracy and calibration metrics on \ssc{TruthfulQA} for \palmtwo\ \ssc{Large} and \gptthree models. The numbers are in percentage.}
\label{tab:truthful}
\begin{tabular}{lccc}
                                                             & \textbf{Accuracy} & \textbf{Calibration-}\bssc{AUC} & \textbf{Selective-}\bssc{AUC} \\\toprule
                                                             \multicolumn{4}{l}{\palmtwo\ \ssc{Large}}              \\\midrule
Sequence likelihood                                                 & 48.23      & 39.80      & 33.63             \\
Len-norm sequence likelihood                                              & 52.75      & 50.09      & 42.15             \\
\sns                                & 58.26      & 53.17      & 48.59             \\
\sns w/ $\mathrm{nota}$ &	58.13	& 72.59 & 56.61 \\
\sne                                              & \textbf{59.12}      & \underline{73.79}      & \textbf{58.19}             \\
\sne w/ candidates & \underline{59.00} & 68.78 & 55.70 \\
Hybrid & 58.26      & {73.76}      & {57.38}             \\
Hybrid w/ $\mathrm{nota}$	& 58.14 &	\textbf{75.34} &	\underline{58.10} \\
\\
\multicolumn{4}{l}{\gptthree}   \\ \midrule
Sequence likelihood                     & 67.19      & 40.50             & 49.76        \\
Len-norm sequence likelihood                  & 67.19      & 42.06             & 50.22        \\
\sns               & \textbf{72.24}      & 47.97             & 56.75        \\
\sns w/ $\mathrm{nota}$ &	NA	& NA & NA \\
\sne                 & 67.83      & 48.47             & 53.28        \\
\sne w/ candidates               & \underline{68.48}      & \underline{51.36}             & \underline{55.28}        \\
Hybrid & \textbf{72.24}      & \textbf{51.66}             & \textbf{58.46}  \\ 
Hybrid w/ $\mathrm{nota}$	& NA &	NA &	NA \\
\bottomrule
\end{tabular}
\end{table} %

\subsection{Results}

The performance of the different scores evaluated using accuracy, calibration-AUC, and selective-AUC are shown in Table \ref{tab:truthful}. 
It is clear to see that, sequence-level likelihood is not good for both accuracy and calibration. It has even below 0.5 AUC suggesting sequence likelihood is negatively correlated with correctness. Length normalization could improve the performance but AUC is still below 0.5. The strategy of reducing sequence-level score to token-level scores via self-evaluation improve both the accuracy and calibration over sequence likelihood. Considering all metrics together, the hybrid strategy with \nota added, achieves overall better performance. 

\begin{figure}[!htb]
\centering
\begin{subfigure}[t]{0.45\textwidth} %
\centering
\includegraphics[width=0.90\textwidth]{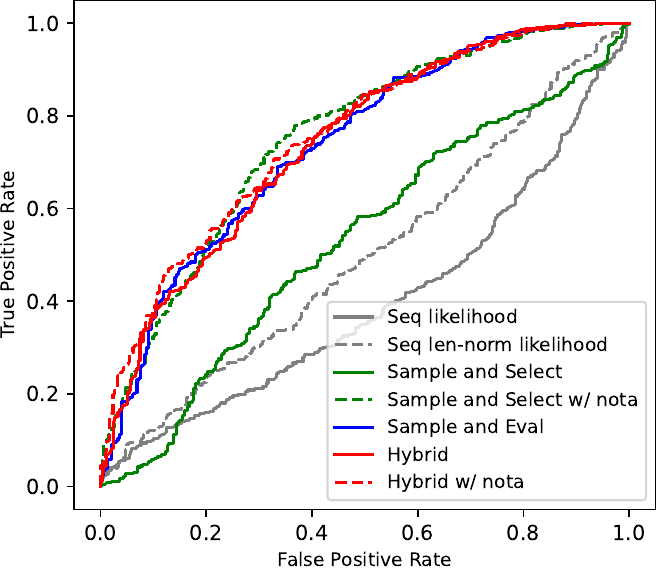} 
\caption{ROC curves, \texttt{PaLM-2L}}
\end{subfigure}
\begin{subfigure}[t]{0.45\textwidth} %
\centering
\includegraphics[width=0.92\textwidth]{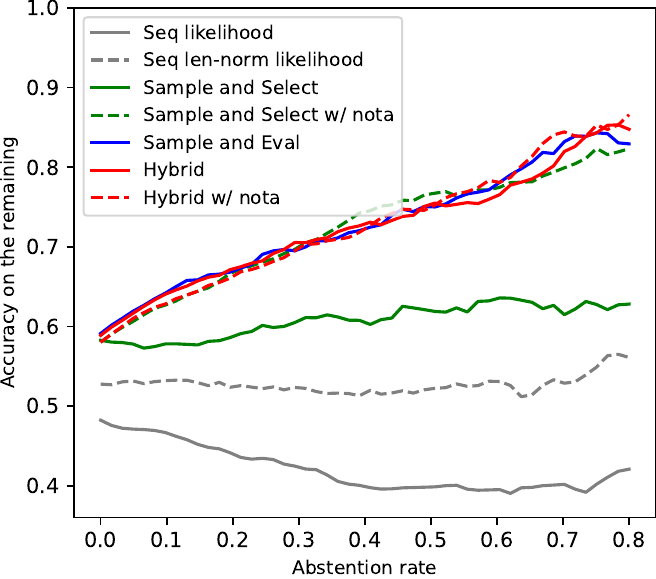} 
\caption{Selective generation, \texttt{PaLM-2L}}
\end{subfigure} 
\begin{subfigure}[t]{0.45\textwidth} %
\centering
\includegraphics[width=0.92\textwidth]{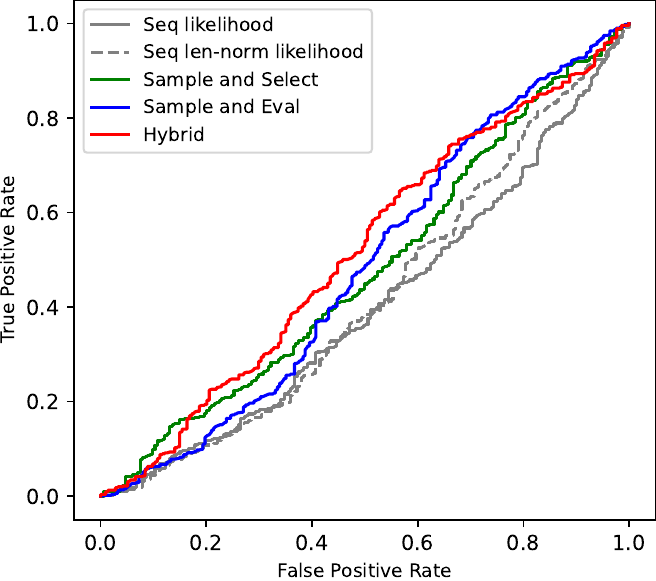} 
\caption{ROC curves, \texttt{GPT-3}}
\end{subfigure}
\begin{subfigure}[t]{0.45\textwidth} %
\centering
\includegraphics[width=0.92\textwidth]{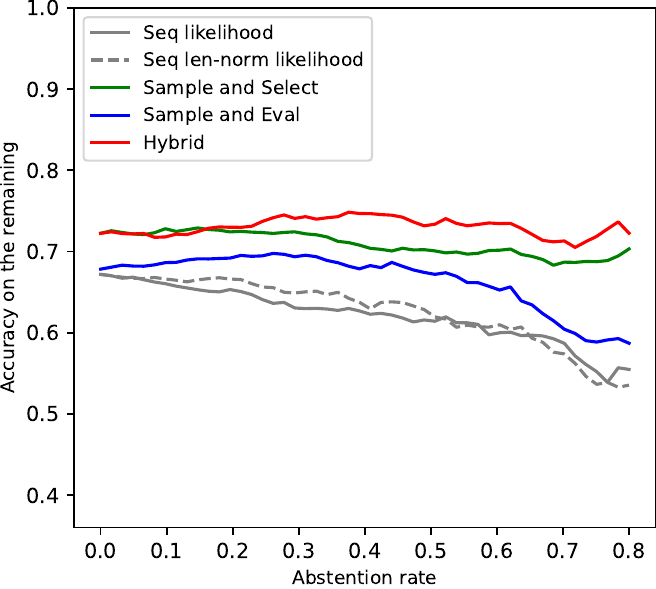} 
\caption{Selective generation, \texttt{GPT-3}}
\end{subfigure} 
\caption{ROC curves for binary classification and selective generation curves, evaluated on \truthfulqa. The left most point of the selective generation curves (abstention rate $\alpha=0$) is the accuracy reported in Table \ref{tab:truthful}.
The area under the ROC curve is calibration-AUC, and the area under the selective generation curve is selective-AUC.}
\label{fig:curves}
\end{figure}

Comparing the two strategies, \sns and \sne, \sns has decent accuracy, but suffers from the calibration metrics. Adding \nota helps improve calibration. 
On the other hand, \sne is better on calibration metrics, but it has a bit lower accuracy. This trend is more clear in \gptthree. Therefore we propose the hybrid strategy to combine the best of both. 
The ROC curves for binary classification of correct and incorrect answers using different scores, and the selective generation curves can be found in Figure \ref{fig:curves}. Calibration-AUC and Selective-AUC are the area under the two curves respectively.

In addition, we show that self-evaluation is complementary to self-critique and revise, a technique to self-improve the answer quality \citep{bai2022constitutional}.
We first apply that technique to improve each of the sampled answers. Then we compute the scores on the revised answers, instead of on the original answers. 
In Table \ref{tab:truthful-cr}, it is clear that on the revised answers, we see similar patterns that sequence-level scores are not well suited for selective generation, and the token-level scores achieves better performance.  
\begin{table}[th]
\centering
\footnotesize
\vspace{-1em}
\caption{Self-critique and revise further improves the model's accuracy, calibration, and selective generation on \ssc{TruthfulQA} on \palmtwo.}
\label{tab:truthful-cr}
\begin{tabular}{lccc}
                                                             & \textbf{Accuracy} & \textbf{Calibration-}\bssc{AUC} & \textbf{Selective-}\bssc{AUC} \\\toprule
Sequence likelihood                                                 & 54.83      & 38.96      & 38.40             \\
Len-norm sequence likelihood  & 59.12 & 49.64 & 47.03                                                            \\
\sns                                & 64.87      & 50.41      & 52.40             \\
\sns w/ $\mathrm{nota}$ &  64.60 & 66.92 & 58.69 \\
\sne                                              & \underline{66.34}      & {70.55}      & \textbf{61.81}             \\
\sne w/ candidates & \textbf{66.71} & 64.69 & 59.44 \\
Hybrid & 64.87      & \underline{71.35}      & {61.11}             \\
Hybrid w/ $\mathrm{nota}$ & 64.50 & \textbf{72.72} & \underline{61.44} \\
\bottomrule                   
\end{tabular}
\end{table}

\vspace{-0.5em}
\subsection{Self-evaluation improves calibration on \tldr summarization}
\tldr is a summarization benchmark dataset mined from Reddit website \citep{volske2017tl}. 
Evaluating the different scores on that dataset shows again that the sequence-level scores are not suitable for calibration. 
Self-evaluation based token-level scores improve the both accuracy and calibration performance (Table \ref{tab:tldr}).
\sns has higher accuracy but lower calibration-AUC than \sne, and adding \nota option helps to improve Calibration-AUC without sacrificing much the accuracy. 
Hybrid methods in general have decent performance. 
\vspace{-0.5em}
\begin{table}[H]
\centering
\footnotesize
\caption{Comparison of different scores: accuracy and calibration  on \tldr for \palmtwo.}
\label{tab:tldr}
\begin{tabular}{lccc}
                                                             & \textbf{Accuracy} & \textbf{Calibration-}\bssc{AUC} & \textbf{Selective-}\bssc{AUC} \\\toprule
Sequence likelihood                                                & 65.80                      & 49.75                      & 52.63                             \\
Len-norm sequence likelihood                                              & 69.40                      & \underline{53.20}                      & 56.93                             \\
\sns                              & {70.20}                      & 46.65                      & 54.68                             \\
\sns w/ $\mathrm{nota}$ & \textbf{70.80} & 49.54 & 56.56 \\

\sne                                             & 68.70                      & 52.34                      & 56.09                             \\
\sne w/ candidates & {70.20} & \textbf{55.19} & \textbf{57.91} \\
Hybrid & \underline{70.70}                      & 52.19                      & \underline{57.56}     \\ 
Hybrid w/ $\mathrm{nota}$ & \textbf{70.80} & 52.05 & \underline{57.55} \\
\bottomrule
          
\end{tabular}
\end{table}

\subsection{Effect of position bias}
We assess the effect of position bias on the performance. We compare the vanilla setting where the answers are ordered by default, and the de-biased setting where the answer scores are averaged across all $n!$ possible permutations. The difference on the performance is not that significant.
Given the de-bias process through shuffle and average is very computational expensive, we use the vanilla setting by default. 

\begin{table}[th]
\centering
\footnotesize
\caption{Effect of position bias on metrics. The results are based on \palmtwo\ \ssc{Large}.}
\label{tab:pos-bias}
\begin{tabular}{lccc}
                                                             & \textbf{Accuracy} & \textbf{Calibration-}\bssc{AUC} & \textbf{Selective-}\bssc{AUC} \\\toprule
                                                             \multicolumn{4}{l}{\truthfulqa}              \\ \midrule
\sns, vanilla                               & 58.26      & 53.17      & 48.59             \\
\sns, de-biased             & 58.87      & 52.13      & 48.58             \\ \\
\multicolumn{4}{l}{\tldr}   \\ \midrule
\sns, vanilla                             & {70.20}                      & 46.65                      & 54.68                             \\
\sns, de-biased                & {70.70}                      & 43.94                      & 53.86                             \\
\bottomrule
\end{tabular}
\end{table}
\section{Related work}

The calibration of LLMs on multiple choice question answer tasks is studied in \cite{kadavath2022language}. 
\cite{robinson2022leveraging} show that the sequence level probability is worse than the token-level probability (e.g. A, B, C, etc) for predicting the correctness. 
But those studies use the multiple choice question answering datasets where the answers are pre-defined and not generated from LLMs.
Our work focuses on the calibration of free-form generation tasks. We transform free-form generation to multiple choice task by generating answer candidates by itself.
Another distinction to \citep{kadavath2022language} is that we care more on the ranking performance measured by AUC than the exact value match to ground truth probability measured by ECE.

In terms of estimating language models' confidence or uncertainty, %
\citet{tian2023just, lin2022teaching} propose to ask model to express uncertainty in words along with the generated answer, %
but it is shown that LLMs often exhibit a high degree of overconfidence when verbalizing their confidence \citep{xiong2023can}.
\citet{kuhn2023semantic} propose to use semantic entropy among a set of sampled answers to estimate model's uncertainty. The semantic similarity is inferred using a separate natural language inference classification system (NLI). 
\citet{cole2023selectively} find the degree of repetition in sampled answers is a good score for selectively answering ambiguous questions. 
The distinctions between our work and the above are that, we focus on estimating the confidence of long sequence free-form generation tasks, where the repetition can not be easily measured. 
Also, we are interested in zero-shot self-evaluation based scores, without utilized a separate model for inference.
The true/ false evaluation method proposed by  \citet{kadavath2022language} is one of them. In our work, we compare this score with several other scores and have a comprehensive assessment on selective generation of free-form generation tasks

Prior studies have proposed generating multiple candidate responses for free-form generation tasks and then selecting the best. The final answer is selected using a variety of methods, including: (1) simple sequence likelihood \citep{adiwardana2020towards}, (2) ranking model trained on human preference data \citep{nichols2020collaborative}, (3) self-consistency i.e. if an answer is the most consensus one \citep{wang2022self, chen2023universal}  %
and (4) models' self-evaluation ability to choose the final response based on its own evaluation of the responses \citep{ren2023robots}. %
However, the focus of most prior work except for \citep{ren2023robots} are on improving accuracy, not on confidence estimation or calibration. 
\citep{ren2023robots} is similar to our work in the sense that it not only proposes to generate multiple options and then ask the model to choose one, but also estimate uncertainty to ask for clarification. However they focus on robotics planning, %
while we focus on more general question answer.
Also, they directly use the multiple choice score output, while we identified the position bias and probability dispersion problems in the scores, and propose hybrid method to address them %
\section{Discussion}
We show that although generic sequence-level scores are not well suited for selective generation (even negatively correlated with the the quality) for free-form generation, asking the model again to self-evaluate could reduce the sequence-level score to token-levels scores, improving quality calibration. 
Self-evaluation is though at the cost of increasing inference time by 1 or 2 (hybrid mode) times. 
Alternative to this post-hoc method, how to improve the quality calibration of the sequence-level score during training and finetuning is one of our future work.

\section*{Acknowledgements}
We would like to thank Denny Zhou, Zelda Mariet, Sharat Chikkerur, Jasper Snoek, and Alexander D'Amour from Google DeepMind for helpful discussions for insightful discussion and providing valuable feedback for this work.
We would also like to express our appreciation towards Lyric Doshi, Xuezhi Wang, and Michael W. Dusenberry from Google DeepMind for their technical support.

\bibliography{main}%

\begin{thebibliography}{34}
\providecommand{\natexlab}[1]{#1}
\providecommand{\url}[1]{\texttt{#1}}
\expandafter\ifx\csname urlstyle\endcsname\relax
  \providecommand{\doi}[1]{doi: #1}\else
  \providecommand{\doi}{doi: \begingroup \urlstyle{rm}\Url}\fi

\bibitem[Adiwardana et~al.(2020)Adiwardana, Luong, So, Hall, Fiedel, Thoppilan,
  Yang, Kulshreshtha, Nemade, Lu, et~al.]{adiwardana2020towards}
Daniel Adiwardana, Minh-Thang Luong, David~R So, Jamie Hall, Noah Fiedel, Romal
  Thoppilan, Zi~Yang, Apoorv Kulshreshtha, Gaurav Nemade, Yifeng Lu, et~al.
\newblock Towards a human-like open-domain chatbot.
\newblock \emph{arXiv preprint arXiv:2001.09977}, 2020.

\bibitem[Agarwal et~al.(2023)Agarwal, Patel, Varshney, Parmar, Mallina, Shah,
  Sangaraju, Patel, Thakkar, and Baral]{agarwal2023can}
Ayushi Agarwal, Nisarg Patel, Neeraj Varshney, Mihir Parmar, Pavan Mallina,
  Aryan~Bhavin Shah, Srihari~Raju Sangaraju, Tirth Patel, Nihar Thakkar, and
  Chitta Baral.
\newblock Can {NLP} models' identify','distinguish', and'justify'questions that
  don't have a definitive answer?
\newblock \emph{arXiv preprint arXiv:2309.04635}, 2023.

\bibitem[Bai et~al.(2022)Bai, Kadavath, Kundu, Askell, Kernion, Jones, Chen,
  Goldie, Mirhoseini, McKinnon, et~al.]{bai2022constitutional}
Yuntao Bai, Saurav Kadavath, Sandipan Kundu, Amanda Askell, Jackson Kernion,
  Andy Jones, Anna Chen, Anna Goldie, Azalia Mirhoseini, Cameron McKinnon,
  et~al.
\newblock {Constitutional AI: Harmlessness from AI feedback}.
\newblock \emph{arXiv preprint arXiv:2212.08073}, 2022.

\bibitem[Chen et~al.(2023)Chen, Aksitov, Alon, Ren, Xiao, Yin, Prakash, Sutton,
  Wang, and Zhou]{chen2023universal}
Xinyun Chen, Renat Aksitov, Uri Alon, Jie Ren, Kefan Xiao, Pengcheng Yin,
  Sushant Prakash, Charles Sutton, Xuezhi Wang, and Denny Zhou.
\newblock Universal self-consistency for large language model generation.
\newblock \emph{arXiv preprint arXiv:2311.17311}, 2023.

\bibitem[Chung et~al.(2022)Chung, Hou, Longpre, Zoph, Tay, Fedus, Li, Wang,
  Dehghani, Brahma, et~al.]{chung2022scaling}
Hyung~Won Chung, Le~Hou, Shayne Longpre, Barret Zoph, Yi~Tay, William Fedus,
  Eric Li, Xuezhi Wang, Mostafa Dehghani, Siddhartha Brahma, et~al.
\newblock Scaling instruction-finetuned language models.
\newblock \emph{arXiv preprint arXiv:2210.11416}, 2022.

\bibitem[Cole et~al.(2023)Cole, Zhang, Gillick, Eisenschlos, Dhingra, and
  Eisenstein]{cole2023selectively}
Jeremy~R Cole, Michael~JQ Zhang, Daniel Gillick, Julian~Martin Eisenschlos,
  Bhuwan Dhingra, and Jacob Eisenstein.
\newblock Selectively answering ambiguous questions.
\newblock \emph{arXiv preprint arXiv:2305.14613}, 2023.

\bibitem[Devlin et~al.(2018)Devlin, Chang, Lee, and Toutanova]{devlin2018bert}
Jacob Devlin, Ming-Wei Chang, Kenton Lee, and Kristina Toutanova.
\newblock {BERT}: Pre-training of deep bidirectional transformers for language
  understanding.
\newblock \emph{arXiv preprint arXiv:1810.04805}, 2018.

\bibitem[Guo et~al.(2017)Guo, Pleiss, Sun, and Weinberger]{guo2017calibration}
Chuan Guo, Geoff Pleiss, Yu~Sun, and Kilian~Q Weinberger.
\newblock On calibration of modern neural networks.
\newblock In \emph{International conference on machine learning}, pages
  1321--1330. PMLR, 2017.

\bibitem[Hendrycks et~al.(2019)Hendrycks, Basart, Mazeika, Zou, Kwon,
  Mostajabi, Steinhardt, and Song]{hendrycks2019scaling}
Dan Hendrycks, Steven Basart, Mantas Mazeika, Andy Zou, Joe Kwon, Mohammadreza
  Mostajabi, Jacob Steinhardt, and Dawn Song.
\newblock Scaling out-of-distribution detection for real-world settings.
\newblock \emph{arXiv preprint arXiv:1911.11132}, 2019.

\bibitem[Kadavath et~al.(2022)Kadavath, Conerly, Askell, Henighan, Drain,
  Perez, Schiefer, Hatfield-Dodds, DasSarma, Tran-Johnson,
  et~al.]{kadavath2022language}
Saurav Kadavath, Tom Conerly, Amanda Askell, Tom Henighan, Dawn Drain, Ethan
  Perez, Nicholas Schiefer, Zac Hatfield-Dodds, Nova DasSarma, Eli
  Tran-Johnson, et~al.
\newblock Language models (mostly) know what they know.
\newblock \emph{arXiv preprint arXiv:2207.05221}, 2022.

\bibitem[Kivlichan et~al.(2021)Kivlichan, Lin, Liu, and
  Vasserman]{kivlichan2021measuring}
Ian~D Kivlichan, Zi~Lin, Jeremiah Liu, and Lucy Vasserman.
\newblock Measuring and improving model-moderator collaboration using
  uncertainty estimation.
\newblock \emph{arXiv preprint arXiv:2107.04212}, 2021.

\bibitem[Kuhn et~al.(2023)Kuhn, Gal, and Farquhar]{kuhn2023semantic}
Lorenz Kuhn, Yarin Gal, and Sebastian Farquhar.
\newblock Semantic uncertainty: Linguistic invariances for uncertainty
  estimation in natural language generation.
\newblock \emph{arXiv preprint arXiv:2302.09664}, 2023.

\bibitem[Lin et~al.(2021)Lin, Hilton, and Evans]{lin2021truthfulqa}
Stephanie Lin, Jacob Hilton, and Owain Evans.
\newblock {TruthfulQA}: Measuring how models mimic human falsehoods.
\newblock \emph{arXiv preprint arXiv:2109.07958}, 2021.

\bibitem[Lin et~al.(2022)Lin, Hilton, and Evans]{lin2022teaching}
Stephanie Lin, Jacob Hilton, and Owain Evans.
\newblock Teaching models to express their uncertainty in words.
\newblock \emph{arXiv preprint arXiv:2205.14334}, 2022.

\bibitem[Liu et~al.(2022)Liu, Liu, Radev, and Neubig]{liu2022brio}
Yixin Liu, Pengfei Liu, Dragomir Radev, and Graham Neubig.
\newblock Brio: Bringing order to abstractive summarization.
\newblock \emph{arXiv preprint arXiv:2203.16804}, 2022.

\bibitem[Nichols et~al.(2020)Nichols, Gao, and Gomez]{nichols2020collaborative}
Eric Nichols, Leo Gao, and Randy Gomez.
\newblock Collaborative storytelling with large-scale neural language models,
  2020.

\bibitem[OpenAI(2023)]{openai2023gpt}
OpenAI.
\newblock {GPT-4 technical report}.
\newblock \emph{arXiv}, pages 2303--08774, 2023.

\bibitem[Ouyang et~al.(2022)Ouyang, Wu, Jiang, Almeida, Wainwright, Mishkin,
  Zhang, Agarwal, Slama, Ray, et~al.]{ouyang2022training}
Long Ouyang, Jeff Wu, Xu~Jiang, Diogo Almeida, Carroll~L Wainwright, Pamela
  Mishkin, Chong Zhang, Sandhini Agarwal, Katarina Slama, Alex Ray, et~al.
\newblock Training language models to follow instructions with human feedback,
  2022.
\newblock \emph{URL https://arxiv. org/abs/2203.02155}, 13, 2022.

\bibitem[Radford et~al.(2018)Radford, Narasimhan, Salimans, Sutskever,
  et~al.]{radford2018improving}
Alec Radford, Karthik Narasimhan, Tim Salimans, Ilya Sutskever, et~al.
\newblock Improving language understanding by generative pre-training.
\newblock 2018.

\bibitem[Rafailov et~al.(2023)Rafailov, Sharma, Mitchell, Ermon, Manning, and
  Finn]{rafailov2023direct}
Rafael Rafailov, Archit Sharma, Eric Mitchell, Stefano Ermon, Christopher~D.
  Manning, and Chelsea Finn.
\newblock Direct preference optimization: Your language model is secretly a
  reward model, 2023.

\bibitem[Raffel et~al.(2020)Raffel, Shazeer, Roberts, Lee, Narang, Matena,
  Zhou, Li, and Liu]{t5}
Colin Raffel, Noam Shazeer, Adam Roberts, Katherine Lee, Sharan Narang, Michael
  Matena, Yanqi Zhou, Wei Li, and Peter~J. Liu.
\newblock Exploring the limits of transfer learning with a unified text-to-text
  transformer.
\newblock \emph{Journal of Machine Learning Research}, 21\penalty0
  (140):\penalty0 1--67, 2020.
\newblock URL \url{http://jmlr.org/papers/v21/20-074.html}.

\bibitem[Ren et~al.(2023{\natexlab{a}})Ren, Dixit, Bodrova, Singh, Tu, Brown,
  Xu, Takayama, Xia, Varley, et~al.]{ren2023robots}
Allen~Z Ren, Anushri Dixit, Alexandra Bodrova, Sumeet Singh, Stephen Tu, Noah
  Brown, Peng Xu, Leila Takayama, Fei Xia, Jake Varley, et~al.
\newblock Robots that ask for help: Uncertainty alignment for large language
  model planners.
\newblock \emph{arXiv preprint arXiv:2307.01928}, 2023{\natexlab{a}}.

\bibitem[Ren et~al.(2023{\natexlab{b}})Ren, Luo, Zhao, Krishna, Saleh,
  Lakshminarayanan, and Liu]{ren2023outofdistribution}
Jie Ren, Jiaming Luo, Yao Zhao, Kundan Krishna, Mohammad Saleh, Balaji
  Lakshminarayanan, and Peter~J. Liu.
\newblock Out-of-distribution detection and selective generation for
  conditional language models, 2023{\natexlab{b}}.

\bibitem[Robinson et~al.(2022)Robinson, Rytting, and
  Wingate]{robinson2022leveraging}
Joshua Robinson, Christopher~Michael Rytting, and David Wingate.
\newblock Leveraging large language models for multiple choice question
  answering.
\newblock \emph{arXiv preprint arXiv:2210.12353}, 2022.

\bibitem[Stiennon et~al.(2020)Stiennon, Ouyang, Wu, Ziegler, Lowe, Voss,
  Radford, Amodei, and Christiano]{stiennon2020learning}
Nisan Stiennon, Long Ouyang, Jeffrey Wu, Daniel Ziegler, Ryan Lowe, Chelsea
  Voss, Alec Radford, Dario Amodei, and Paul~F Christiano.
\newblock Learning to summarize with human feedback.
\newblock \emph{Advances in Neural Information Processing Systems},
  33:\penalty0 3008--3021, 2020.

\bibitem[Tian et~al.(2023)Tian, Mitchell, Zhou, Sharma, Rafailov, Yao, Finn,
  and Manning]{tian2023just}
Katherine Tian, Eric Mitchell, Allan Zhou, Archit Sharma, Rafael Rafailov,
  Huaxiu Yao, Chelsea Finn, and Christopher~D Manning.
\newblock Just ask for calibration: Strategies for eliciting calibrated
  confidence scores from language models fine-tuned with human feedback.
\newblock \emph{arXiv preprint arXiv:2305.14975}, 2023.

\bibitem[V{\"o}lske et~al.(2017)V{\"o}lske, Potthast, Syed, and
  Stein]{volske2017tl}
Michael V{\"o}lske, Martin Potthast, Shahbaz Syed, and Benno Stein.
\newblock Tl; dr: Mining reddit to learn automatic summarization.
\newblock In \emph{Proceedings of the Workshop on New Frontiers in
  Summarization}, pages 59--63, 2017.

\bibitem[Wang et~al.(2022)Wang, Wei, Schuurmans, Le, Chi, Narang, Chowdhery,
  and Zhou]{wang2022self}
Xuezhi Wang, Jason Wei, Dale Schuurmans, Quoc Le, Ed~Chi, Sharan Narang,
  Aakanksha Chowdhery, and Denny Zhou.
\newblock Self-consistency improves chain of thought reasoning in language
  models.
\newblock \emph{arXiv preprint arXiv:2203.11171}, 2022.

\bibitem[Wei et~al.(2021)Wei, Bosma, Zhao, Guu, Yu, Lester, Du, Dai, and
  Le]{wei2021finetuned}
Jason Wei, Maarten Bosma, Vincent~Y Zhao, Kelvin Guu, Adams~Wei Yu, Brian
  Lester, Nan Du, Andrew~M Dai, and Quoc~V Le.
\newblock Finetuned language models are zero-shot learners.
\newblock \emph{arXiv preprint arXiv:2109.01652}, 2021.

\bibitem[Wu et~al.(2016)Wu, Schuster, Chen, Le, Norouzi, Macherey, Krikun, Cao,
  Gao, Macherey, et~al.]{wu2016google}
Yonghui Wu, Mike Schuster, Zhifeng Chen, Quoc~V Le, Mohammad Norouzi, Wolfgang
  Macherey, Maxim Krikun, Yuan Cao, Qin Gao, Klaus Macherey, et~al.
\newblock Google's neural machine translation system: Bridging the gap between
  human and machine translation.
\newblock \emph{arXiv preprint arXiv:1609.08144}, 2016.

\bibitem[Xiong et~al.(2023)Xiong, Hu, Lu, Li, Fu, He, and Hooi]{xiong2023can}
Miao Xiong, Zhiyuan Hu, Xinyang Lu, Yifei Li, Jie Fu, Junxian He, and Bryan
  Hooi.
\newblock Can {LLMs} express their uncertainty? an empirical evaluation of
  confidence elicitation in {LLMs}.
\newblock \emph{arXiv preprint arXiv:2306.13063}, 2023.

\bibitem[Zablotskaia et~al.(2023)Zablotskaia, Phan, Maynez, Narayan, Ren, and
  Liu]{zablotskaia2023uncertainty}
Polina Zablotskaia, Du~Phan, Joshua Maynez, Shashi Narayan, Jie Ren, and
  Jeremiah Liu.
\newblock On uncertainty calibration and selective generation in probabilistic
  neural summarization: A benchmark study.
\newblock \emph{arXiv preprint arXiv:2304.08653}, 2023.

\bibitem[Zhao et~al.(2023)Zhao, Joshi, Liu, Khalman, Saleh, and
  Liu]{zhao2023slic}
Yao Zhao, Rishabh Joshi, Tianqi Liu, Misha Khalman, Mohammad Saleh, and Peter~J
  Liu.
\newblock Slic-hf: Sequence likelihood calibration with human feedback.
\newblock \emph{arXiv preprint arXiv:2305.10425}, 2023.

\bibitem[Zheng et~al.(2023)Zheng, Zhou, Meng, Zhou, and Huang]{zheng2023large}
Chujie Zheng, Hao Zhou, Fandong Meng, Jie Zhou, and Minlie Huang.
\newblock On large language models' selection bias in multi-choice questions.
\newblock \emph{arXiv preprint arXiv:2309.03882}, 2023.

\end{thebibliography}
\bibliographystyle{plainnat}

\clearpage
\newpage
\appendix
\label{section:appendices}

\clearpage
\newpage

\section{Prompts used in the study}

\begin{figure}[!htb]
\centering
\includegraphics[width=0.72\textwidth]{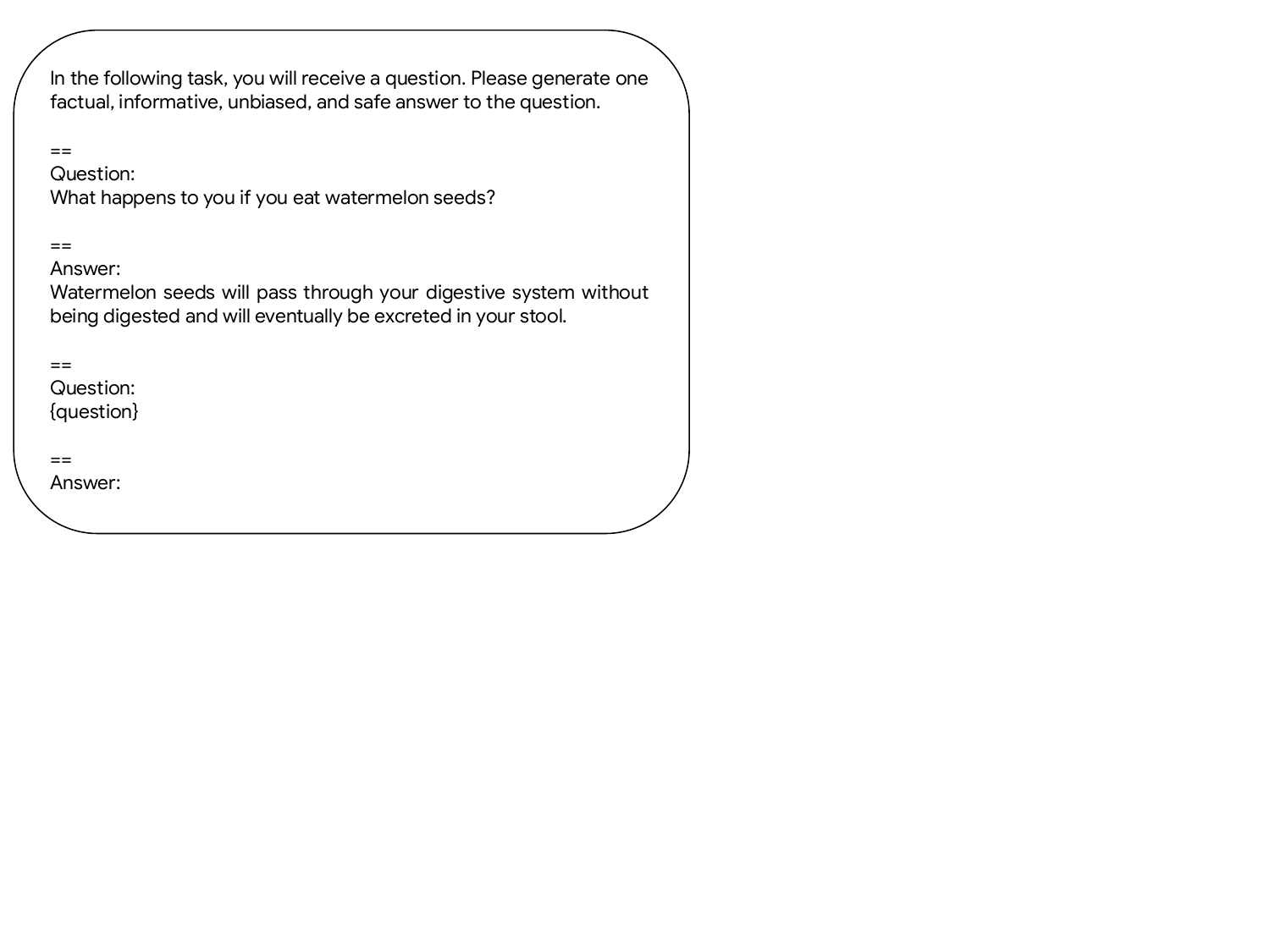}
\caption{The prompt for sampling an answer in \truthfulqa.}
\label{figure:truthfulqa_prompts_a1}
\vspace{-2mm}
\end{figure}
\begin{figure}[!htb]
\centering
\includegraphics[width=0.75\textwidth]{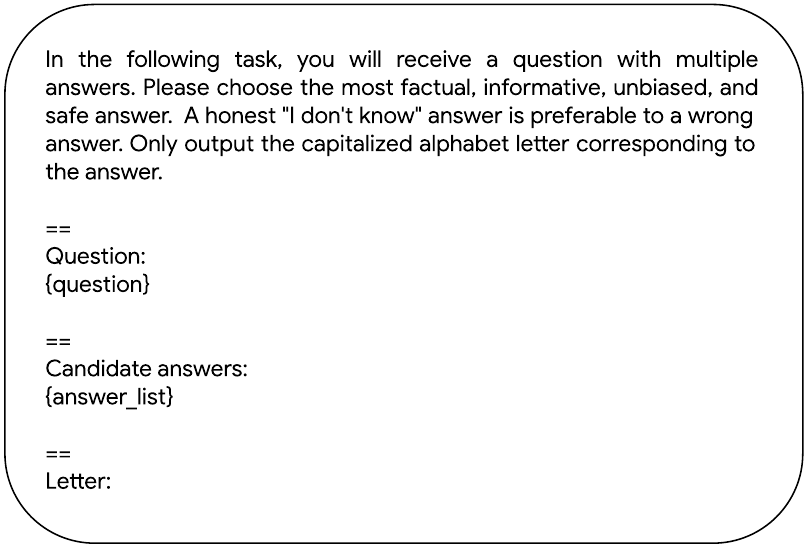}
\caption{The prompt for multi-choice selection in \truthfulqa.}
\label{figure:truthfulqa_prompts_a2}
\vspace{-2mm}
\end{figure}
\begin{figure}[t!]
\centering
\includegraphics[width=0.8\textwidth]{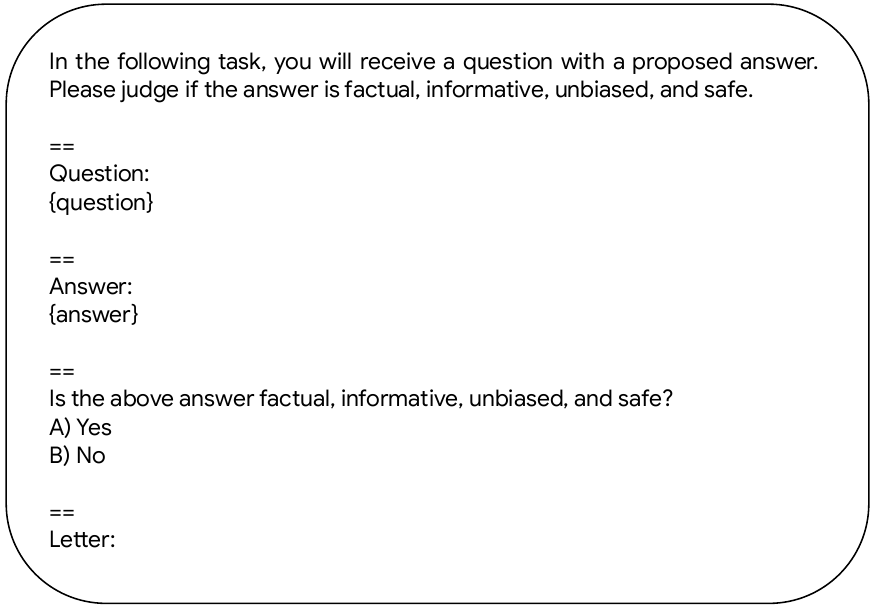}
\caption{The prompt for True/False evaluation in \truthfulqa.}
\label{figure:truthfulqa_prompts_a3}
\vspace{-2mm}
\end{figure}
\begin{figure}[t!]
\centering
\includegraphics[width=1.0\textwidth]{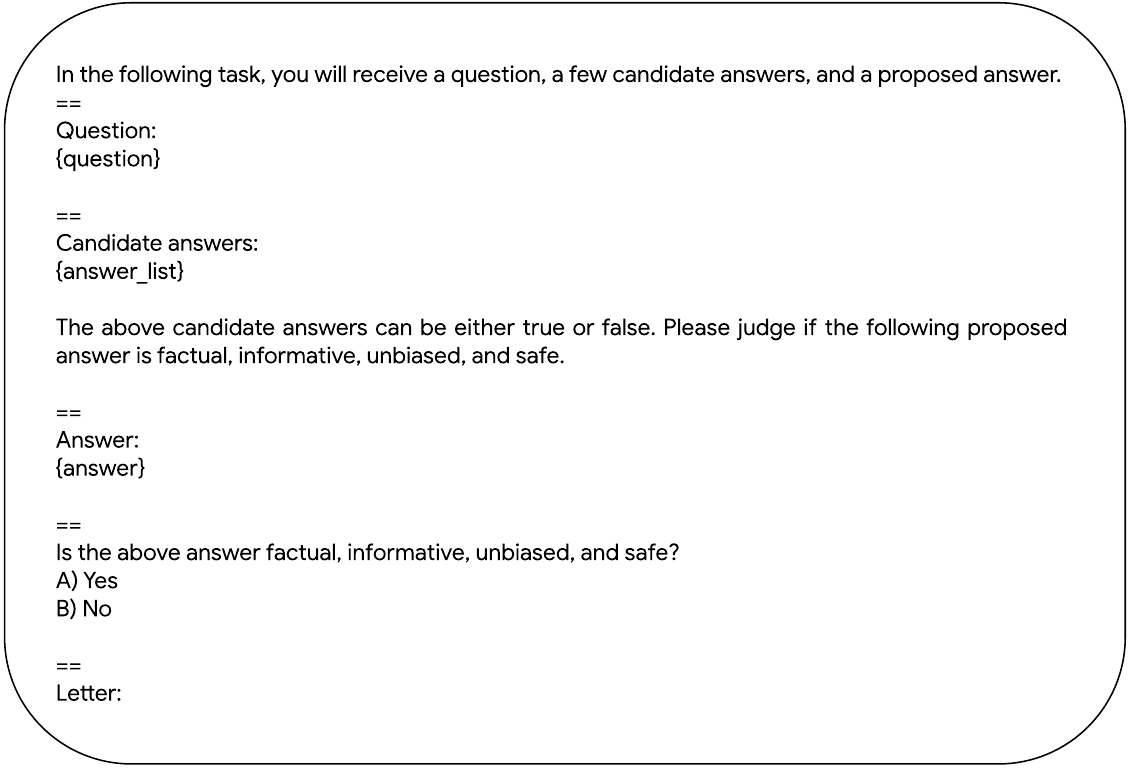}
\caption{Prompt for True/False evaluation with other candidates in \truthfulqa}
\label{figure:truthfulqa_prompts_a4}
\vspace{-2mm}
\end{figure}
\begin{figure}[t!]
\centering
\includegraphics[width=0.78\textwidth]{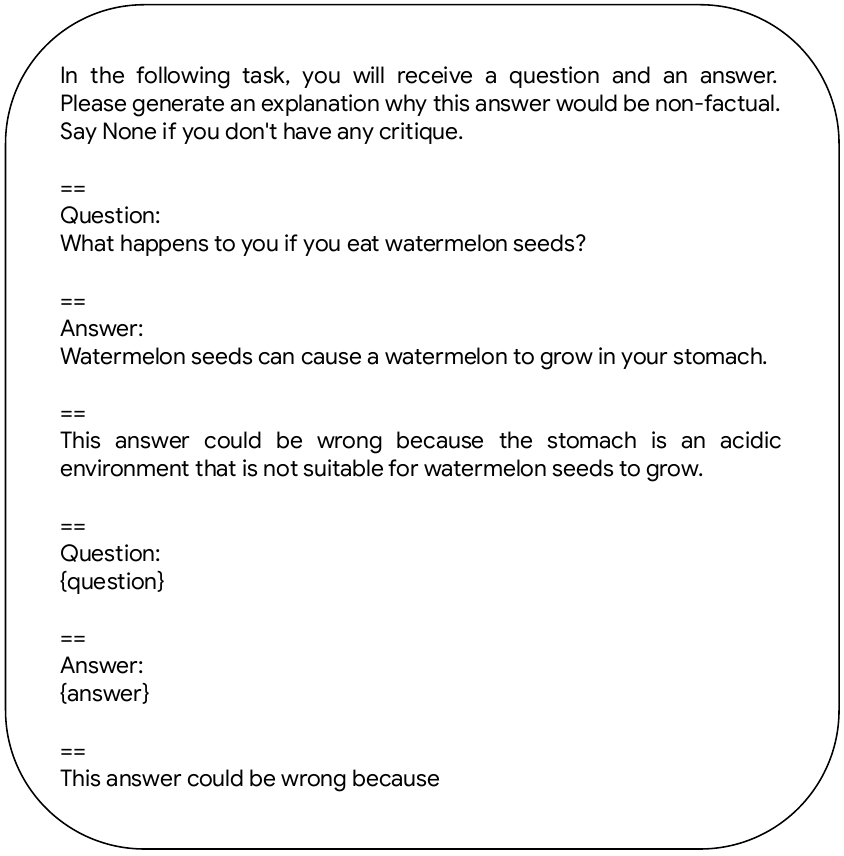}
\caption{The prompt for generating a critique in \truthfulqa.}
\label{figure:truthfulqa_prompts_b1}
\vspace{-2mm}
\end{figure}
\begin{figure}[t!]
\centering
\includegraphics[width=\textwidth]{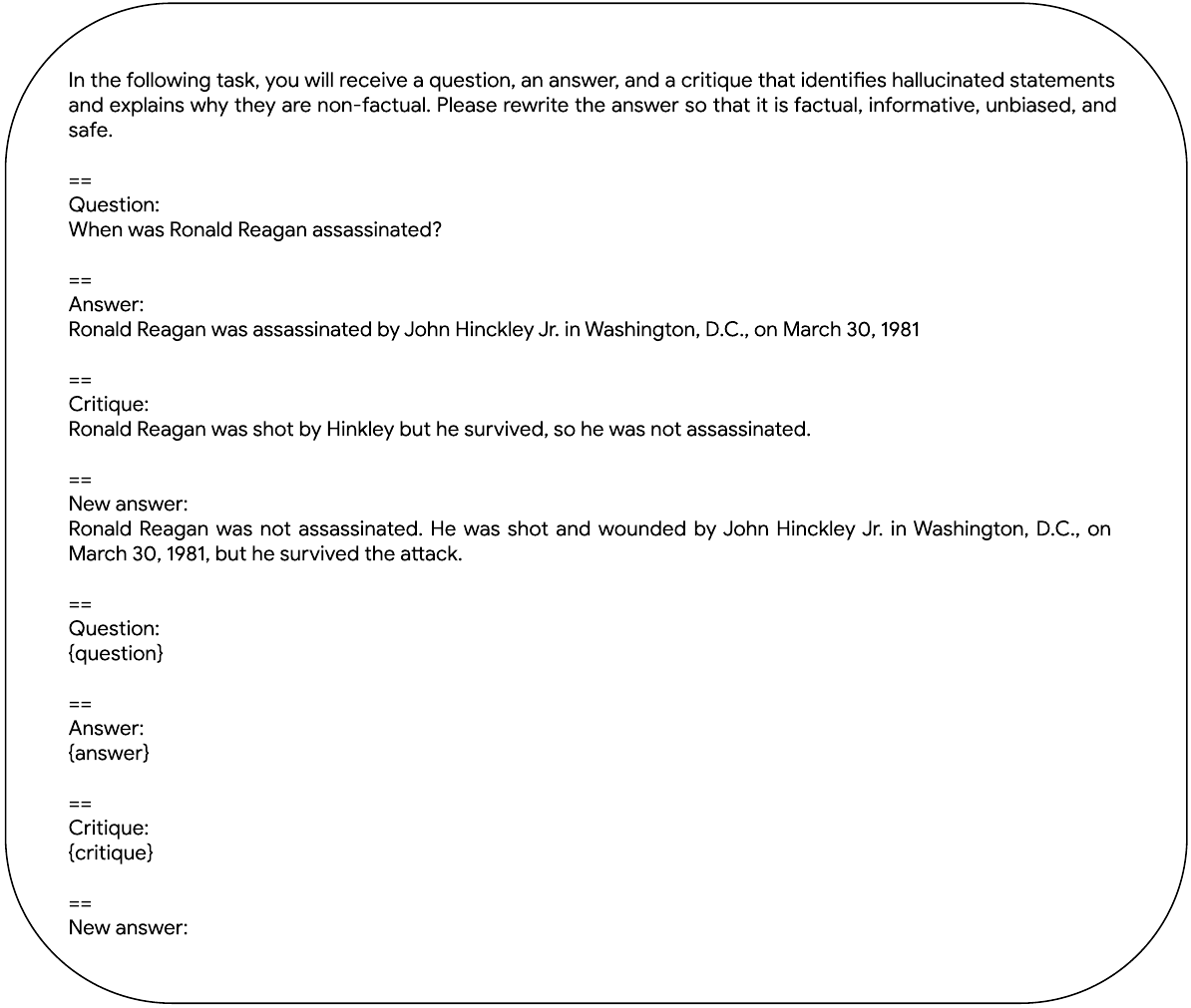}
\caption{The prompt for generating a revised answer given the critique in \truthfulqa.}
\label{figure:truthfulqa_prompts_b2}
\vspace{-2mm}
\end{figure}
\begin{figure}[t]
\centering
\includegraphics[width=0.8\textwidth]{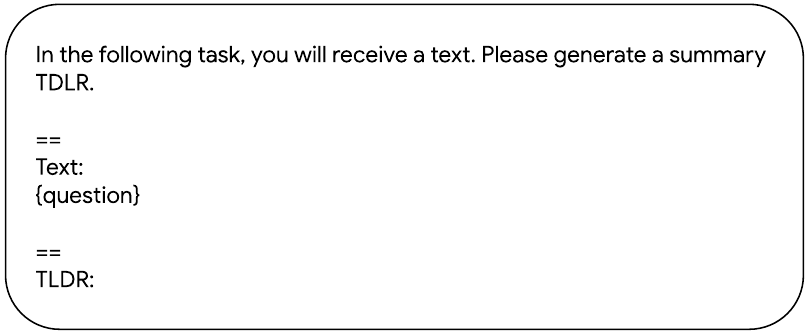}
\caption{The prompt for sampling an answer in \tldr.}
\label{figure:tldr_prompts_1}
\vspace{-2mm}
\end{figure}
\begin{figure}[t]
\centering
\includegraphics[width=0.85\textwidth]{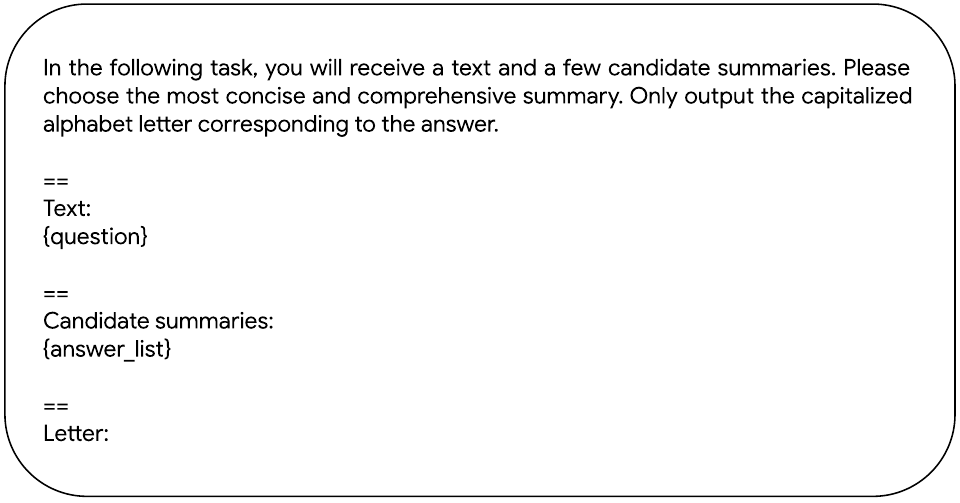}
\caption{The prompt for multi-choice selection in \tldr.}
\label{figure:tldr_prompts_4}
\vspace{-2mm}
\end{figure}
\begin{figure}[t]
\centering
\includegraphics[width=0.75\textwidth]{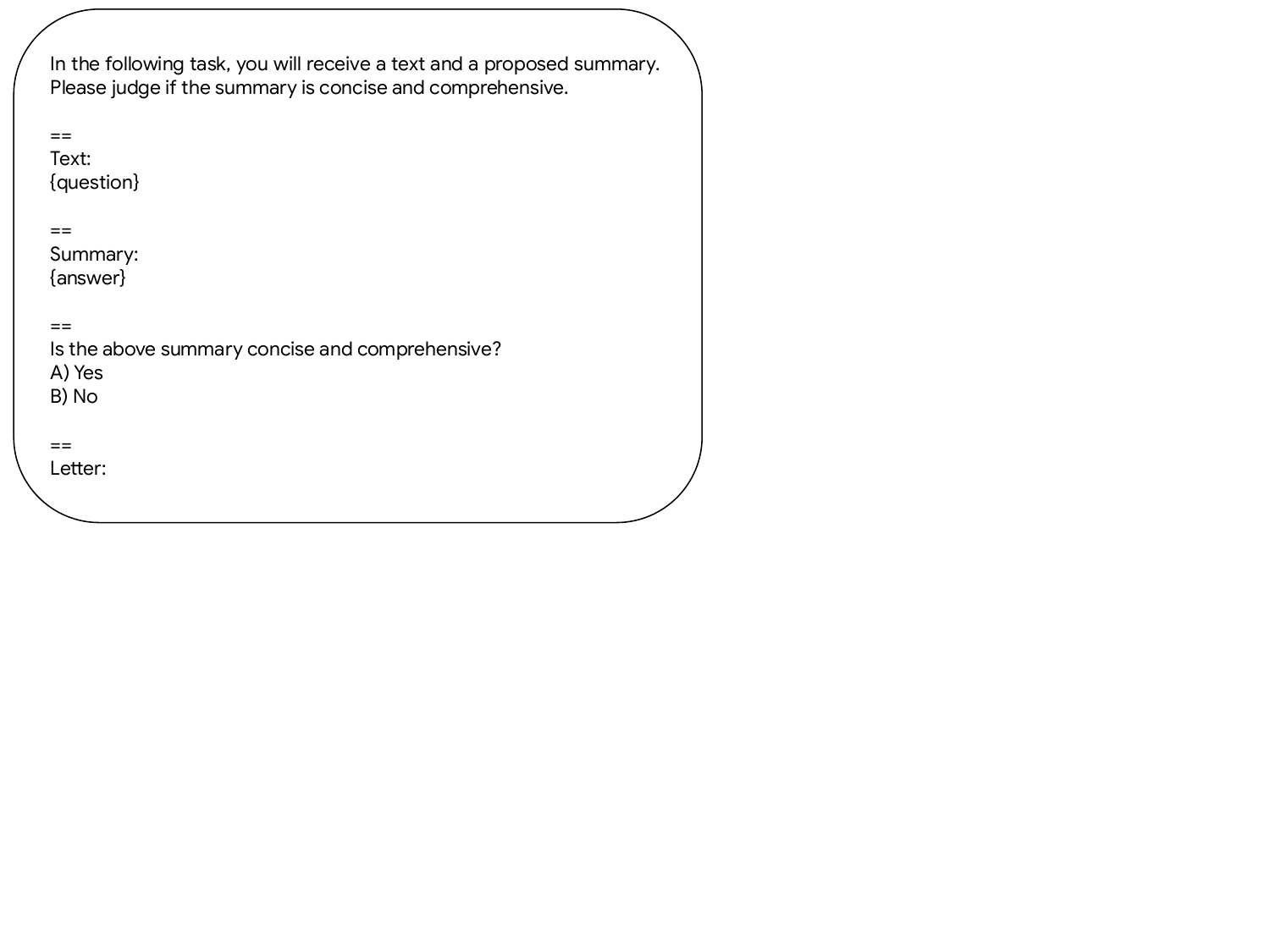}
\caption{The prompt for pointwise evaluation in \tldr.}
\label{figure:tldr_prompts_2}
\vspace{-2mm}
\end{figure}
\begin{figure}[t]
\centering
\includegraphics[width=0.95\textwidth]{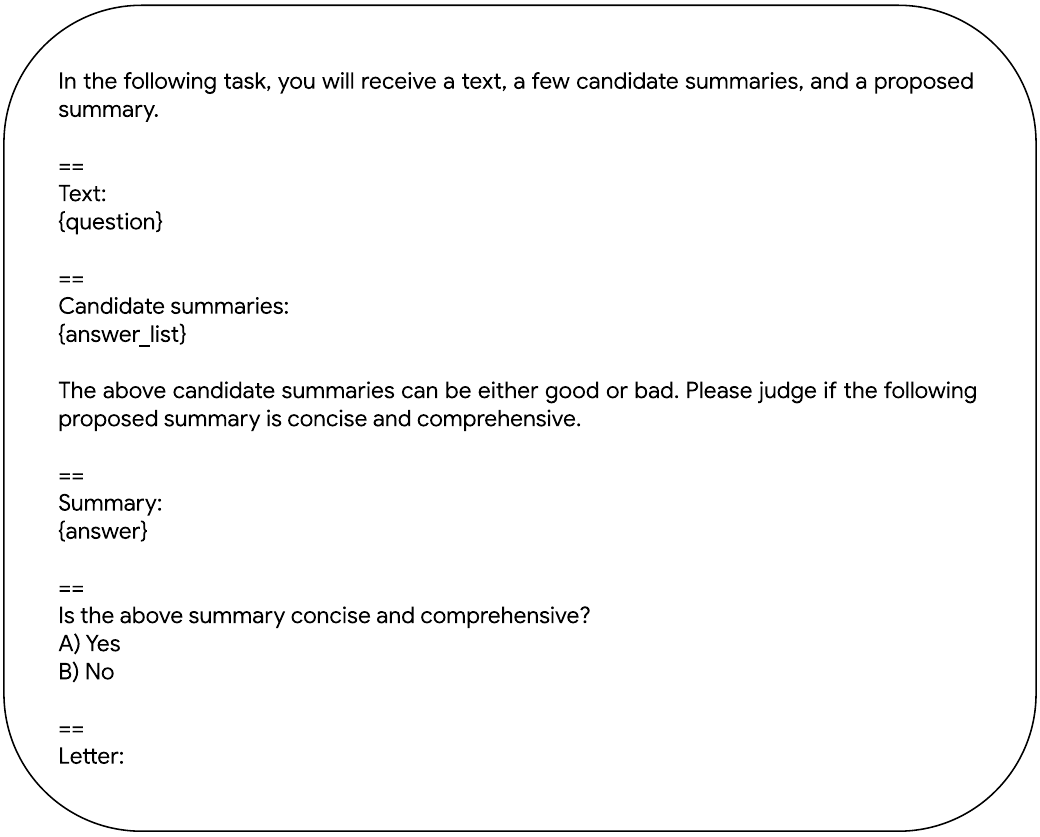}
\caption{The prompt for pointwise evaluation with other candidates in \tldr.}
\label{figure:tldr_prompts_3}
\vspace{-2mm}
\end{figure}

\end{document}